\documentclass{article}
\usepackage{ijcai20}

\usepackage{times}
\usepackage{soul}
\usepackage{url}
\usepackage[hidelinks]{hyperref}
\usepackage[utf8]{inputenc}
\usepackage[small]{caption}
\usepackage{graphicx}
\usepackage{amsmath}
\usepackage{amsthm}
\usepackage{amsfonts}
\usepackage{booktabs}
\usepackage{algorithm}
\usepackage{algorithmic}
\title{Unsupervised Neural Aspect Search with Related Terms Extraction}

\author{
Timur Sokhin$^1$
\and
Maria Khodorchenko$^{1}$\And
Nikolay Butakov$^1$
\affiliations
$^1$ITMO University, St. Petersburg, Russia\\
\emails
qwinpin@gmail.com, mkhodorchenko@niuitmo.ru,
alipoov.nb@gmail.com
}

\begin{document}

\maketitle

\begin{abstract}
  The tasks of aspect identification and term extraction remain challenging in natural language processing. While supervised methods achieve high scores, it is hard to use them in real-world applications due to the lack of labelled datasets. Unsupervised approaches outperform these methods on several tasks, but it is still a challenge to extract both an aspect and a corresponding term, particularly in the multi-aspect setting. In this work, we present a novel unsupervised neural network with convolutional multi-attention mechanism, that allows extracting pairs (aspect, term) simultaneously, and demonstrate the effectiveness on the real-world dataset. We apply a special loss aimed to improve the quality of multi-aspect extraction. The experimental study demonstrates, what with this loss we increase the precision not only on this joint setting but also on aspect prediction only.
\end{abstract}

\section{Introduction}
Unsupervised aspect extraction is an essential part of natural language processing and usually solved using topic modelling approaches, which have proven themselves in this task. In general, aspect extraction aims to identify the category or multiple categories of a given text. The aspect can be a global context of the sentence or a specific term in this sentence; a term, in turn, can be either a single word or a collocation. Previous unsupervised approaches achieved significant improvement in the task of aspect extraction. The joint task of the aspect and the aspect term pairs extraction is still a challenge for natural language processing. For example: in the sentence "Best Pastrami I ever had and great portion without being ridiculous" the aspect and aspect term pairs "Food Quality: Pastrami" and "Food Style option: portion".

Most of the existing approaches apply two-stage extraction: aspect extraction first and then aspect term extraction based on the known aspect. We propose a conjoint solution based on the convolutional multi-attention mechanism (CMAM). The CMAM was inspired by Inception-block in computer vision, where kernels of different sizes allow incorporating features from different levels of localisation. The sentence representations built with CMAM capture the features, which are used for aspect predictions, while the attention detects related terms. Also, the convolutional attention does not require much additional time to infer the result, which is vital for the real-world application. In order to increase the quality of multi-aspect extraction, we propose a novel loss function - triplet-like aspect spreading (TLAS), which maximises the distance between top-N aspect-based sentence representations and minimises the distance between these representations and corresponding aspect vectors. This approach allows achieving close to the state-of-the-art results in aspect extraction with the ability to extract their terms.
In summary, the contributions of this paper are:
\begin{itemize}
    \item CMAM; convolutional multi-attention mechanism, which is aimed to build sentence vector representation and to extract aspect terms.
    \item TLAS -  loss function for aspect probabilities distribution modifying.
    \item The experimental study of the proposed model on SemEval-2016 Restaurant dataset and Citysearch corpus.
\end{itemize}

\section{Related Work}
Aspect extraction is applied in different tasks: sentiment analysis, documents categorisation. In 2014 the task of aspect-based sentiment analysis was started, and significant results have been achieved both in aspect extraction and aspect term extraction in different languages \cite{pontiki-etal-2016-semeval,Yan2013,Wang2015,He2017,Luo2019}. However, usually, a topic is a general category, which describes the whole document or the sentence. This is a severe problem in case if we need to know more details. Also, some texts contain multiple topics, which is detected by topic modelling, but they are not possible to identify which part of the document is responsible for a particular topic.

The simplest way to solve this problem - two-stage extraction. Firstly we detect aspect, then extract term using this knowledge \cite{cetin-etal-2016-tgb,brun-etal-2016-xrce,Toh2016}. We are interested in the system that incorporates both capabilities in its core. Also, these approaches are supervised; this limits their applicability in the real world.

\section{The proposal}
\begin{figure}[t]
\includegraphics[width=8cm]{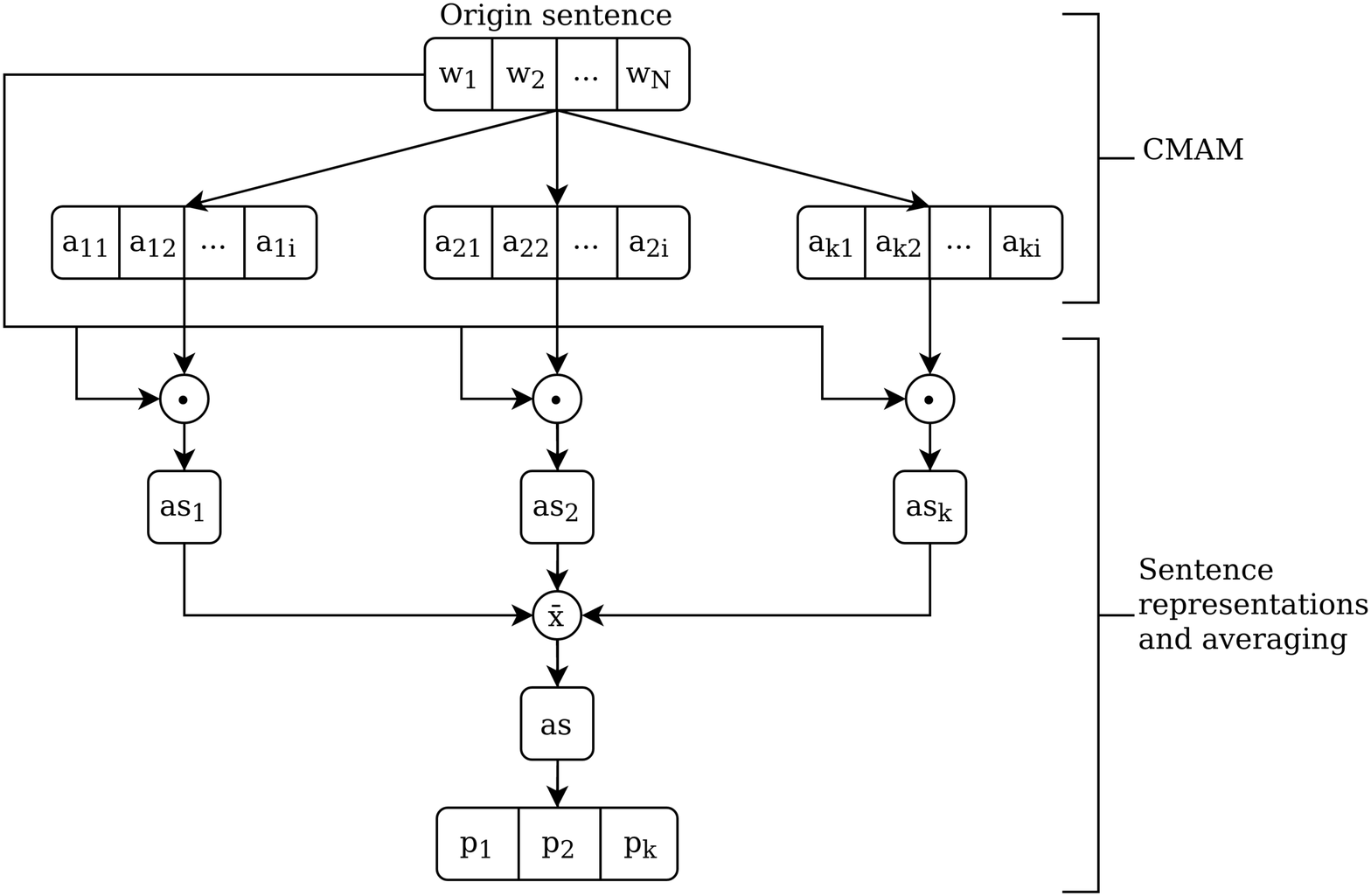}
\caption{Schema of the proposed approach to aspect and aspect term extraction.}
\centering
\label{fig:Approach}
\end{figure}
The general neural attention model for aspect extraction is based on ABAE idea \cite{He2017}, which significantly outperformed previous methods of unsupervised aspect predictions. In this model, the set of aspects is learned in the same embedding space with words and can be easily interpreted by a human. However, despite the similarities at the level of aspect matrix generation, our approach differs at the core of attention mechanism: we aim to capture the most relevant information from the sentence for each aspect encountered independently - Figure~\ref{fig:Approach}. The main idea can be shown as:
\begin{enumerate}
    \item Build attention for each aspect using CMAM.
    \item Build sentence representations for each of these attentions - attentioned sentence representations (AS\textsubscript{j}).
    \item Build attentioned averaged sentence (AS) representation with AS\textsubscript{j} averaging.
    \item Infer the weights of aspects.
\end{enumerate}
After we have the weights of aspects, we can select the aspect term(-s) from corresponding attentions.

\subsection{Model}
\subsubsection{Convolutional multi-attention mechanism}
CMAM makes it possible to extract term for each individual aspect using simple convolutional approach. The set of convolutions with different kernel sizes allows analysing each word of the sentence on different levels to take into account the local and the global context.
\begin{figure}[t]
\includegraphics[width=8cm]{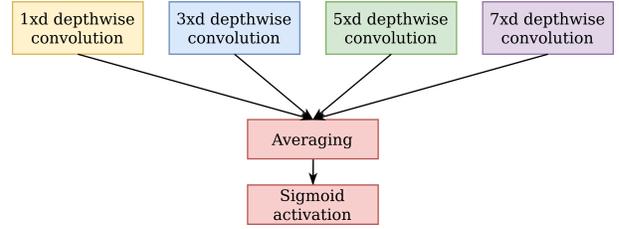}
\caption{CMAM architecture. d in the kernel size denotes the embedding dimension.}
\centering
\label{fig:CMAM}
\end{figure}
CMAM consists of F depthwise convolutions Conv\textsubscript{j} with K channels - Figure~\ref{fig:CMAM}:
\begin{equation}
    {A}' = \frac{1}{F}\sum_{j=1}^{F}S \ast Conv_j
\end{equation}
\begin{equation}
    A = \frac{1}{1 + e^{-{A}'}}
\end{equation}
\begin{equation}
    AS_j = A_j \cdot S
\end{equation}
\begin{equation}\label{eq:AS}
    AS = \frac{1}{K}\sum_{j=1}^{K}AS_j
\end{equation}
where \textit{S} \( \in \mathbb{R}^{N \times d} \) - sentence, represented as a set of N word embeddings w\textsubscript{i} with dimension d; \textit{A} \( \in \mathbb{R}^{N \times K}  \) is an attention matrix with \textit{K} - number of aspects. \textit{AS}\textsubscript{j} is attentioned sentence representation - the summary "meaning" over a given aspect.

\subsubsection{Aspect prediction}
Averaging in Eq.~\ref{eq:AS} is used for the aspect probabilities prediction using simple linear transformation (Eq.~\ref{eq:app}).
\begin{equation}\label{eq:app}
    p = \frac{1}{1 + e^{-(wAS + b)}}
\end{equation}{}
where w - learnable weights, b - learnable bias.

We use Sigmoid activation function for the reason that a sentence can contain more than one aspect and the Softmax are not applicable here.

\subsubsection{Sentence reconstruction}
Sentence reconstruction is required for the model training in an unsupervised way. Like an auto-encoder, we match the reconstructed sentence \textit{RS} with the with averaged embeddings of the original sentence. The reconstructed sentence, in our case, is the linear combination of \textit{AS} and aspect embedding matrix \textit{AEM} - a weighted sum of all aspects.

\subsubsection{Aspect and aspect term inferring}
According to Eq.~\ref{eq:app} for each aspect we have weight. The rule by which the aspects describing the sentence are selected: we calculate the q-th quantile \textit{q}\textsubscript{as} of all weights and use it as the minimum threshold. \textit{N}\textsubscript{as} of the remaining aspects with the highest weight is selected.

For attentions of these aspect the same rule is applied: q-th quantile of all words weight \textit{q}\textsubscript{at} and \textit{N}\textsubscript{at} of the words are above the threshold is selected as a terms. \textit{q}\textsubscript{as}, \textit{N}\textsubscript{as}, \textit{q}\textsubscript{at} and \textit{N}\textsubscript{at} are hyper-parameters of the model.

\subsection{Training objective}
Our approach consist of three loss functions:
\begin{itemize}
    \item Hinge loss \textit{H}, which demonstrates better result comparing with Triplet loss. We minimise the reconstruction error in the form of the inner product between \textit{RS} and averaged embeddings of original sentence \textit{S} and maximise the difference between \textit{RS} and randomly sampled negative sentences, which are formed as an averaging of words embeddings per sample.
    \item Orthogonality regularisation aimed to make the aspect embeddings more representative and unique.
    \item Triplet-like aspect spreading - since we want to be able to detect more than one aspect, TLAS makes top-N aspects more unique, helps to avoid repetitions of similar aspects.
\end{itemize}
\subsubsection{Orthogonality loss}
Orthogonal loss in Eq.~\ref{eq:ORT} allows us to make the vectors of individual aspects more unique. However, our experiments demonstrate that, when orthogonality value is close to zero, aspects can "degenerate" and cover only a small area of a particular category. This effect leads to an increase in the quality of determination of the aspects themselves but reduces the overall efficiency of pair extraction <aspect: aspect term>.
\begin{equation}\label{eq:ORT}
    U = \lambda \left( \left \|AEM \cdot AEM^T - I\right \| - s \right)
\end{equation}
where \textit{s} is an offset, which determines to which value \textit{U} will be optimised, \( \lambda \) is a weight of this loss, \textit{I} - identity matrix.

\begin{table}
\centering
\begin{tabular}{llr}  
\toprule
New label  & Origin label & Size \\
\midrule
Ambience       & Ambience General  & 82     \\
\midrule
Food        & Food Quality  & 447      \\
            & Food Style options      \\
            & Drinks Quality      \\
            & Drinks Style options      \\
\midrule
Staff     & Service General & 157 \\
\midrule
Multi-labels    & All samples with  & 337 \\
                & more than one label \\
\midrule
Prices (Omitted)     & Drinks Price  & 15      \\
            & Food Price      \\
            & Restaurant Price      \\
\midrule
Location (Omitted)  & Location General & 8 \\
\midrule
Restaurant (Omitted)    & Restaurant General & 144 \\
\midrule
Misc (Omitted)  & Restaurant Miscellaneous & 32 \\
\bottomrule
\end{tabular}
\caption{Map of origin labels and the new. "Omitted" denotes categories, which are not used for evaluation.}
\label{tab:map_o2n}
\end{table}

\subsubsection{TLAS}
The idea is that the top-N aspect of each sentence should cover different categories:
\begin{equation}
    \sqrt{(AS_j - AS_l)^2} \to \infty
\end{equation}
where \textit{j} and \textit{l} are indexes of the top-2 aspects. We select the number 2, because of the mean number of aspects equal to 2.25 among all sentences with more than 1 aspect labelled by a human reviewer.

From the other hand, each top-2 \textit{AS}\textsubscript{j} must be closer to corresponding aspect vector from aspect embedding matrix - \textit{AEM}\textsubscript{j}. The overall TLAS loss is formulated as:
\begin{align}
    T =  & max(0, 1 + \sqrt{\sum_{k=1}^{d}(AS_{jk} - AEM_{jk})^2} - \nonumber\\
    -   & \sqrt{\sum_{k=1}^{d}(AS_{jk} - AS_{lk})^2})
\end{align}
Summary loss function \textit{L} is:
\begin{equation}
    L = H + U + T
\end{equation}

\section{Experiments}
\subsection{Experimental settings}
\begin{table}
\centering
\begin{tabular}{lll}  
\toprule
Generated  & Representative Words & Gold-label \\
aspect\\
\midrule
Drinks        & cocktail, liquor, beer  & Food     \\
Ingredients       & ceviche, oxtail, saffron     \\
Misc. Food       & biscuit, onion, bun     \\
Menu       & menu, selection, option     \\
Cuisine style       & food, cuisine, fusion     \\
Geography       & mexican chinese japanese     \\
of cuisine\\
\midrule
Staff       & manager, maitre, politely  & Staff     \\
Staff       & service, waitstaff, staff     \\
Attitude       & attentive, friendly, polite      \\
\midrule
Environment       & couch, fireplace, patio   & Ambience     \\
Style       & retro, deco, overhead     \\
\midrule
Price       & overpriced, average   & Price     \\
Location       & manhattan, brooklyn, ny   & Misc.     \\
Opinion terms       & great, fantastic, amazing   & Misc     \\
Place          & place restaurant spot & Restaurant\\

\bottomrule
\end{tabular}
\caption{Inferred aspects, their representative words and gold-labels mapping. Price and misc. gold-labels were omitted due to ambiguity of their meanings.}
\label{tab:aspect2gold}
\end{table}

\subsubsection{Datasets}
Out experiments conducted as unconstrained SemEval task: we use the whole Citysearch corpus to train our model and to evaluate aspect extraction efficiency and use SemEval-2016 Restaurant reviews dataset \cite{pontiki-etal-2016-semeval} to evaluate the aspect and aspect term co-extraction efficiency.

The joint aspect and aspect term extraction was a part of task 5 "Aspect Based Sentiment Analysis". Due to under-representation or ambiguity of some categories, the final evaluation labels presented in the Table~\ref{tab:map_o2n}. Restaurant category was omitted because of the ambiguity of terms selected by SemEval reviewers - 45.8\% of all terms for this aspect are "place" and "restaurant". All samples in these categories contain one pair (aspect: aspect term). Samples with more than one pair are grouped in the multi-label category.

\subsubsection{Experimental settings}
We use word2vec model trained on Citysearch reviews with embedding size 200, window size 10 and negative size 20 for embedding matrix initialisation. Texts filtered from stop words, punctuation. Total vocabulary size is 9000 that covers 77.1\% of SemEval dataset. Words out of vocabulary marked as "unknown", and if they were predicted, we consider them as an error. While in works related to unsupervised aspect extraction \cite{He2017,Zhao2010,Brody2010} the optimal number of aspects is 14 for the Restaurant domain, our experiments demonstrate that the best result is achieved using 30 pre-defined aspects initialised with centroids of k-means clustering over embedding matrix. Model is trained for 5 epochs using Adam optimiser: learning rate 0.0005, default betas 0.9 and 0.999; with batch size 64. Orthogonal loss weight is set to 0.5, offset 0.3. 
\subsubsection{Evaluation settings}
\begin{figure}
\includegraphics[width=8cm,height=7cm]{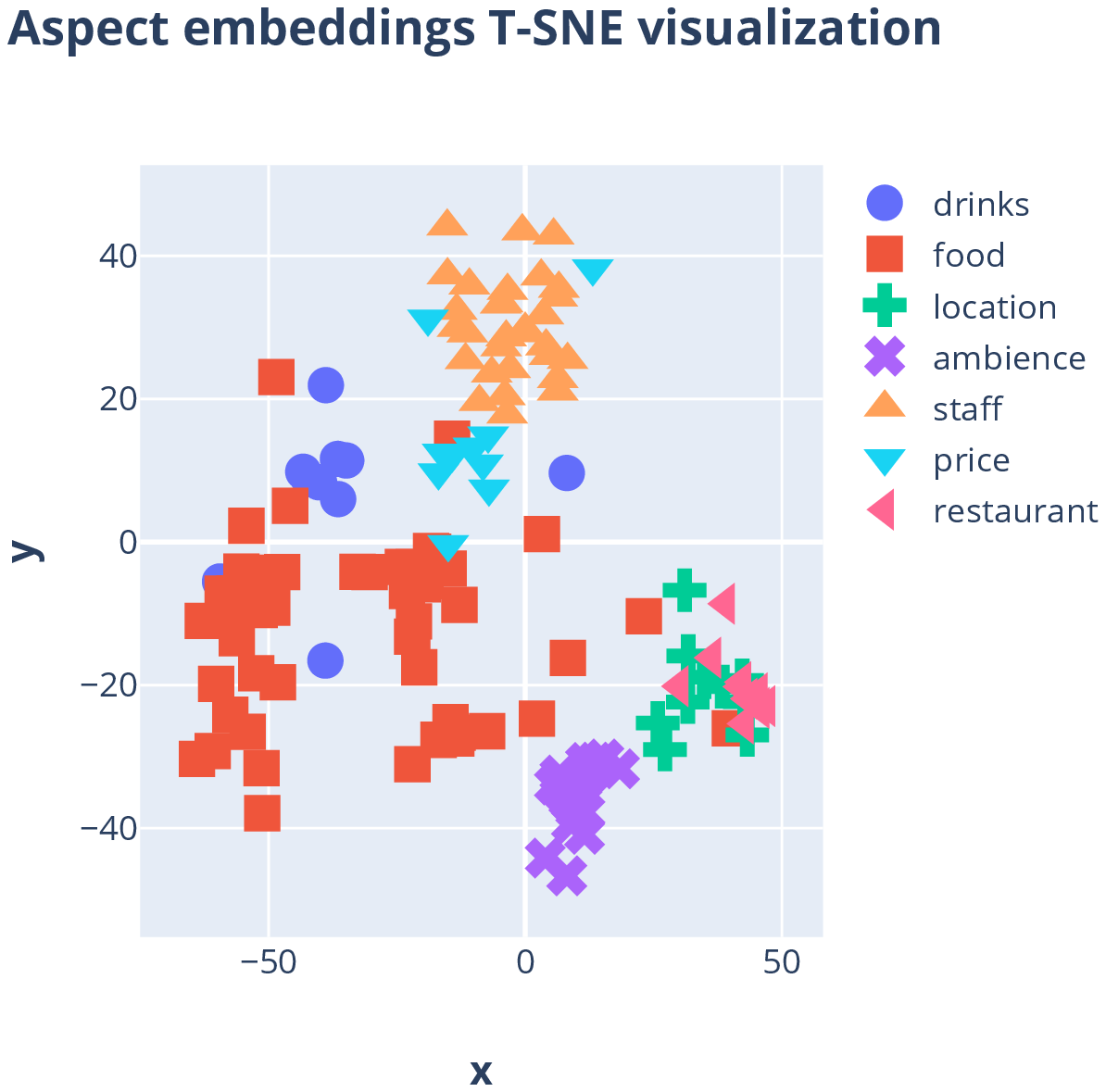}
\caption{Visualisation of aspects embeddings using T-SNE.}
\centering
\label{fig:asp_tsne}
\end{figure}

The nearest words can describe the generated aspect embeddings in embeddings space. For the task of aspect extraction, each of 30 generated aspects was mapped with one of the six gold-standard label - Table~\ref{tab:aspect2gold}. T-SNE visualisation of generated aspect is shown in Figure~\ref{fig:asp_tsne}. Restaurant and location topics located in one area, which confirms the ambiguity of these aspects. Food and drinks aspects also overlap and are located next to the cluster of price aspects. That is in line with the nature of these topics. The partial coincidence of the predicted term and gold-term with the correct aspect is also considered as the true positive. That is also done due to the ambiguity of data labels.

\subsection{Aspect extraction results}
Topic (aspect) modelling approaches such as LDA-based \cite{Blei2003} retain their significance in this task, and they can still compete with modern methods. Moreover, new methods are being developed that improve the flexibility of the probabilistic approach - ARTM \cite{Vorontsov2015}. The results of BTM \cite{Yan2013}, SERBM \cite{Wang2015} are taken from \cite{He2017}. Approaches based on neural networks are also actively developing: ABAE, AE-CSA \cite{He2017,Luo2019}.

Compared to the others, our approach demonstrates close to state-of-the-art results - Table~\ref{tab:aspect_results}. Also, we analyse the result of our model without TLAS loss. In spite of rather high efficiency, the decrease is noticeable both in precision and in recall. In addition, \(\lambda\)=0.0 in Eq.~\ref{eq:ORT} leads to even greater reduction of F1-score despite high precision value. Further, we demonstrate that in the joint problem, that leads to more significant deterioration.

\begin{figure*}[t]
\centering
\includegraphics[width=\textwidth, width=16cm]{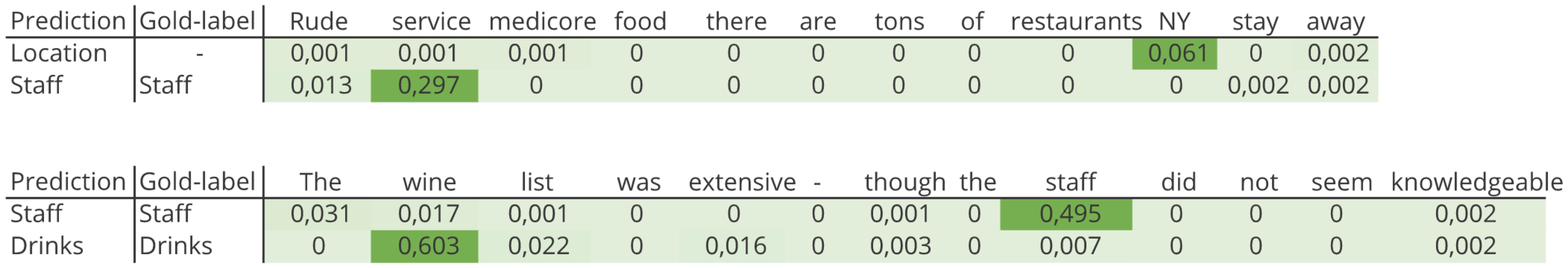}
\caption{Examples of aspect and attention extraction.}
\label{fig:example}
\end{figure*}

\begin{table}[t]
\centering
\begin{tabular}{llrrr}  
\toprule
Aspect  & Model & Precision & Recall & F1 \\
\midrule
Food        &  BTM  &  0.933  &  0.745  &  0.816\\
        &  SERBM  &  0.891  &  0.854  &  0.872\\
        &  ABAE  &  0.953  &  0.741  &  0.828\\
        &  AE-CSA  &   0.903  &  0.926  &    0.914\\
\cmidrule(lr){2-5}
        &  Ours  &  0.887  &  0.945  &  0.915\\
        &  Ours without &  0.910  &  0.890  &  0.900\\
        &  TLAS\\
        &  Ours, \(\lambda\)=0.0 &  0.817  &  0.909  &  0.860\\
\midrule
Staff        &  BTM  &  0.828  &   0.579  &  0.677\\
        &  SERBM  &   0.819  &  0.582  &  0.680\\
        &  ABAE  &  0.802  &  0.728  &  0.757\\
        &  AE-CSA  &   0.804  &  0.756  &   0.779\\
\cmidrule(lr){2-5}
        &  Ours  &  0.804  &  0.676  &  0.735\\
        &  Ours without &  0.668  &  0.662  &  0.665\\
        &  TLAS\\
        &  Ours, \(\lambda\)=0.0 &  0.852  &  0.554  &  0.671\\
\midrule
Ambience        &  BTM  &  0.813  &  0.599  &  0.685\\
        &  SERBM  &  0.805  &  0.592  &   0.682\\
        &  ABAE  &  0.815  &  0.698  &   0.740\\
        &  AE-CSA  &  0.768  &  0.773  &   0.770\\
\cmidrule(lr){2-5}
        &  Ours  &  0.763  &  0.757  &  0.760\\
        &  Ours without &  0.672  &  0.733  &  0.701\\
        &  TLAS\\
        &  Ours, \(\lambda\)=0.0 &  0.817  &  0.677  &  0.741\\
\bottomrule
\end{tabular}
\caption{Aspect extraction results.}
\label{tab:aspect_results}
\end{table}

\subsection{Aspect and aspect term extraction results}
\begin{table}[b]
\centering
\begin{tabular}{lrrr}  
\toprule
Aspect  & Ours (F1) & without TLAS & \(\lambda\)=0.0 (F1) \\
        &           & (F1)\\
\midrule
Ambience    & 0.279 & 0.157  & 0.227\\
Food        & 0.426 &  0.196  & 0.349\\
Staff       & 0.501 &  0.165  & 0.406\\
Multi-label    & 0.296 &  0.103  & 0.173\\
Micro- & 0.369 &  0.149  & 0.272\\
average\\
\bottomrule
\end{tabular}
\caption{F1 score for aspect and aspect term extraction for different model variations.}
\label{tab:aspect_term_res}
\end{table}
We provide the experimental results for different aspects in Table~\ref{tab:aspect_term_res}. We could not find strong dependency between quality of aspect detection and term extraction. We can see the influence of TLAS loss and orthogonality shift especially at the multi-label subtask. In Figure~\ref{fig:example} we provide two examples of sentences and their aspect and aspect term. The drawbacks of our approach is the strict definition of the term: non-aspect terms weights are almost zero. Solving this problem will allow you to more accurately handle the search for phrases. Also, we must note the difficulty of evaluating of unsupervised approaches on SemEval-2016 restaurant dataset. Supervised approaches work well with strict rules and control (labels); unsupervised is not limited by outside and can give result, which is not obvious for a human reviewer at first sight.

\section{Conclusions}
In this paper, we presented a new unsupervised neural model with the convolutional multi-attention mechanism for aspect search with related term extraction. In the experimental study, we show the efficiency of our model in the task of aspect extraction compared to the other state-of-the-art approaches and its possibility of correct aspect and aspect term joint identification.

The further work can be directed to the enhancement of the inferring procedure to incorporate multiple aspects with high predicted probability within the same category and their terms and also to modification of attention mechanism in order to achieve more smooth weights distribution between words of the sentence.

\section*{Acknowledgements}
This work financially supported by Ministry of Education and Science of the
Russian Federation, Agreement \#x.x.x.x (xx/xx/xx). Unique Identification xxx.

\bibliographystyle{named}
\bibliography{ijcai20}

\end{document}